\documentclass[final]{cvpr}

\usepackage{times}
\usepackage{epsfig}
\usepackage{graphicx}
\usepackage{amsmath}
\usepackage{amssymb}
\usepackage{multirow}
\usepackage{pifont}
\usepackage{booktabs}
\usepackage{epsfig}
\usepackage{booktabs,multirow}
\usepackage{graphics}
\usepackage{threeparttable}
\usepackage{color}
\usepackage[normalem]{ulem}
\usepackage{multirow}
\usepackage{float}
\usepackage{amsfonts}
\usepackage{bm}
\usepackage{enumitem}
\usepackage[numbers]{natbib}
\usepackage{array}
\usepackage[table]{xcolor}
\usepackage{colortbl}
\definecolor{myy}{RGB}{126,95,0}
\definecolor{mygray}{gray}{.9}
\definecolor{bblue}{RGB}{30,80,120}
\definecolor{mygray1}{gray}{.7}
\usepackage{bm}
\usepackage{mathtools}
\usepackage{subcaption}
\usepackage{siunitx}
\usepackage[nopar]{lipsum}

\newcolumntype{I}{!{\vrule width 1pt}}

\DeclareMathOperator*{\argmin}{argmin}

\definecolor{ggray}{RGB}{127,127,127}

\makeatletter
\newcommand{\thickhline}{%
	\noalign {\ifnum 0=`}\fi \hrule height 1pt
	\futurelet \reserved@a \@xhline
}
\makeatother

\usepackage{caption}
\usepackage[pagebackref=true,breaklinks=true,colorlinks,bookmarks=false]{hyperref}
\usepackage[utf8]{inputenc}

\usepackage{cleveref}
\crefname{section}{§}{§§}
\Crefname{section}{§}{§§}

\def\cvprPaperID{6630} 
\def\confYear{CVPR 2021}

\begin{document}

\title{Differentiable Multi-Granularity Human Representation Learning for \\ Instance-Aware Human Semantic Parsing}

\author{Tianfei Zhou$^{1}$~, Wenguan Wang$^{1}$\thanks{Corresponding author: \textit{Wenguan Wang}.}~~, Si Liu$^2$~, Yi Yang$^3$~, Luc Van Gool$^{1}$\\
\small{$^1$Computer Vision Lab, ETH Zurich~~~$^2$Institute of Artificial Intelligence, Beihang University~~~$^3$University of Technology Sydney}\\
\small\url{https://github.com/tfzhou/MG-HumanParsing}
}

\maketitle
\thispagestyle{empty}

\begin{abstract}
	To address the challenging task of instance-aware human part parsing, a new bottom-up regime is proposed to learn category-level human semantic segmentation as well as multi-person pose estimation in a joint and end-to-end manner. It is a compact, efficient and powerful framework that exploits structural information over different human granularities and eases the difficulty of person partitioning. Specifically, a dense-to-sparse projection field, which allows explicitly associating dense human semantics with sparse keypoints, is learnt and progressively improved over the network feature pyramid for robustness. Then, the difficult pixel grouping problem is cast as an easier, multi-person joint assembling task. By formulating joint association as maximum-weight bipartite matching, a differentiable solution is developed to exploit projected gradient descent and Dykstra's cyclic projection algorithm. This makes our method end-to-end trainable and allows back-propagating the grouping error to directly supervise multi-granularity human representation learning. This is distinguished from current bottom-up human parsers or pose estimators which require sophisticated post-processing or heuristic greedy algorithms. Experiments on three instance-aware human parsing datasets show that our model outperforms other bottom-up alternatives with much more efficient inference.
\end{abstract}

\begin{figure}[t]
	\begin{center}
		\includegraphics[width=\linewidth]{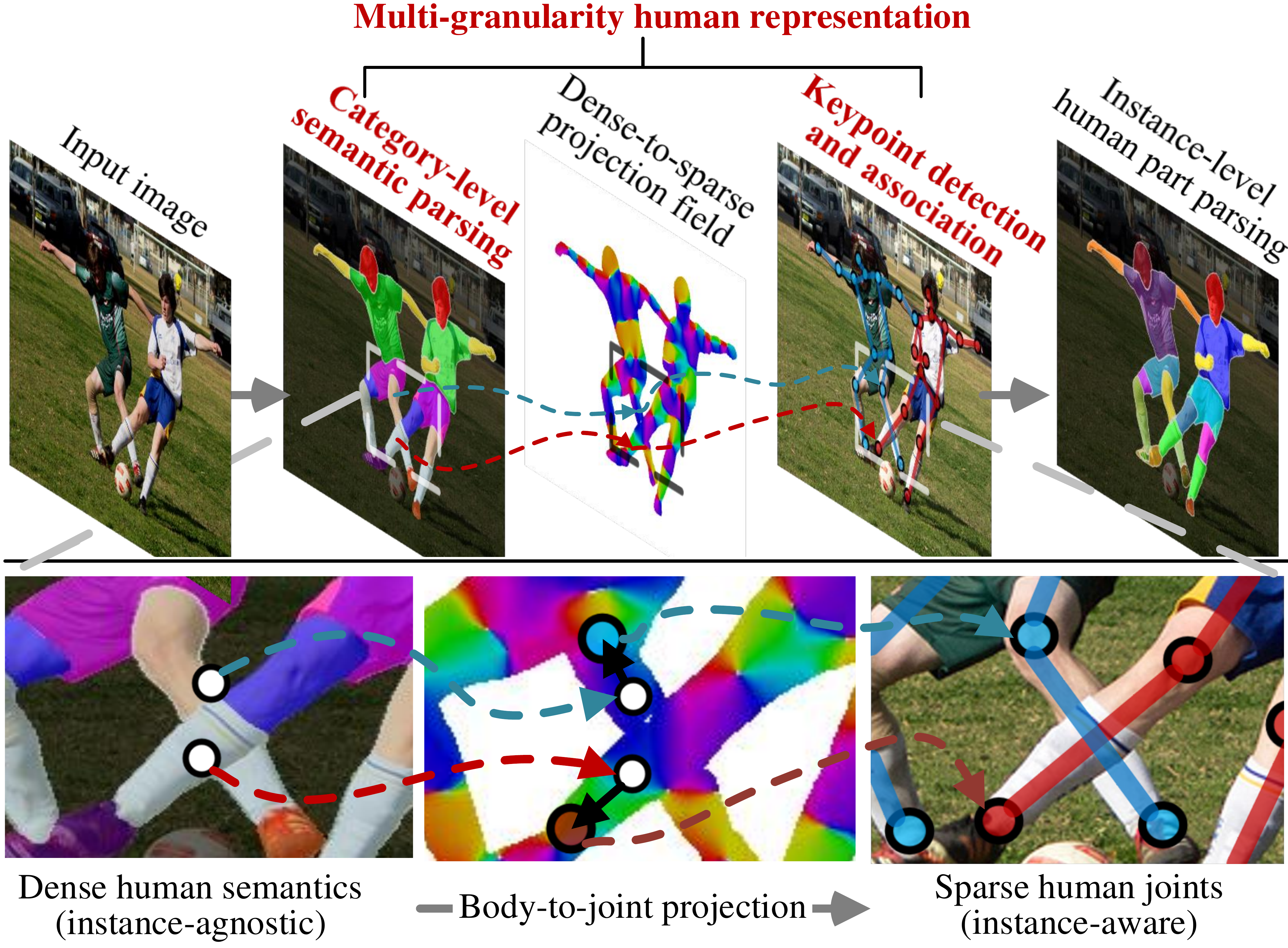}
	\end{center}
	\vspace{-15pt}
	\captionsetup{font=small}
	\caption{\small{\textbf{Overview of our new bottom-up regime for instance-aware human semantic parsing.} By learning \textbf{1)} \textit{category-level} human semantic parsing, \textbf{2)} \textit{body-to-joint} projection, and  \textbf{3)} \textit{bottom-up} keypoint detection and association in a joint and end-to-end manner, our model tackles the task in a differentiable,  multi-granularity human representation learning framework.}}
	\label{fig:motivation}
	\vspace{-15pt}
\end{figure}

\section{Introduction}
Instance-aware$_{\!}$ human$_{\!}$ parsing, \ie, partitioning$_{\!}$ humans into semantic parts\!~(\eg, torso, head) and associating each part with the corresponding human instance, has only started to be tackled in the literature (dating back to\!~\cite{li2017holistic}). This article addresses this task through a new regime, which learns to jointly estimate human poses (\ie, a sparse, skeleton-based human representation) and segment human parts (\ie, a pixel-wise, fine-grained human representation) in an \textit{end-to-end trainable}, \textit{bottom-up} fashion.

In the field of human parsing, the idea of leveraging human pose as structural knowledge to facilitate human understanding has been exploited for years~\cite{yamaguchi2012parsing}. However, previous efforts only focus on the \textit{instance-agnostic} setting~\cite{xia2016pose,xia2017joint,gong2017look,nie2018mutual,zhang2020correlating,wang2021hierarchical}. Further, most of them directly utilize human joints (pre-detected from off-the-shelf pose estimators) as extra inputs~\cite{yamaguchi2012parsing,xia2016pose}, or simply generate key-point estimations as a by-product~\cite{gong2017look,zhang2020correlating}. In sharp contrast, by learning to associate semantic person pixels with their closest
person instance joints, our model seamlessly injects bottom-up pose estimation into instance-aware human semantic learning and inference.  Thus, our human parser can make use of the complementary cues from sparse human joints and dense part semantics, and push further the envelope of human understanding in unconstrained environments. This represents an early effort to formulate instance-aware human parsing and multi-person pose estimation in a bottom-up, differentiable, and multi-granularity human representation learning framework (see~\figref{fig:motivation}).

More importantly, our framework yields a new paradigm for$_{\!}$ bottom-up, instance-aware$_{\!}$ human$_{\!}$ part$_{\!}$ parsing.  For$_{\!}$~human instance discrimination, current bottom-up human parsers learn to associate pixels by directly regressing instance locations~\cite{zhao2018understanding} or predicting pair-wise connectiveness between pixels~\cite{li2017multiple}. However, instance locations (\ie, human centroids or bounding box corners) are less semantic and pair-wise embeddings are difficult to learn (due to deformations, variations and occlusions of human bodies)~\cite{wei2020point}.  As shown in~\figref{fig:motivation}, our model instead regresses semantic- and geometry-rich human joints as pixel embeddings. Then, fine-grained human semantics can be efficiently grouped through body joint association. In essence, it  formulates instance-aware human semantic parsing by jointly learning: \textbf{1)} \textit{category-level} human semantic parsing,  \textbf{2)} \textit{body-to-joint} projection, and  \textbf{3)} \textit{bottom-up} multi-person keypoint detection and association. Thus, our model avoids sophisticated inference and heavy network designs, and neatly explores the structural constraints over human bodies. Such a flexible network design can benefit from progress in bottom-up pose estimation and semantic segmentation techniques, and significantly differentiates itself from existing instance-aware human parsers.

%


\begin{figure}[t]
	\begin{center}
		\includegraphics[width=\linewidth]{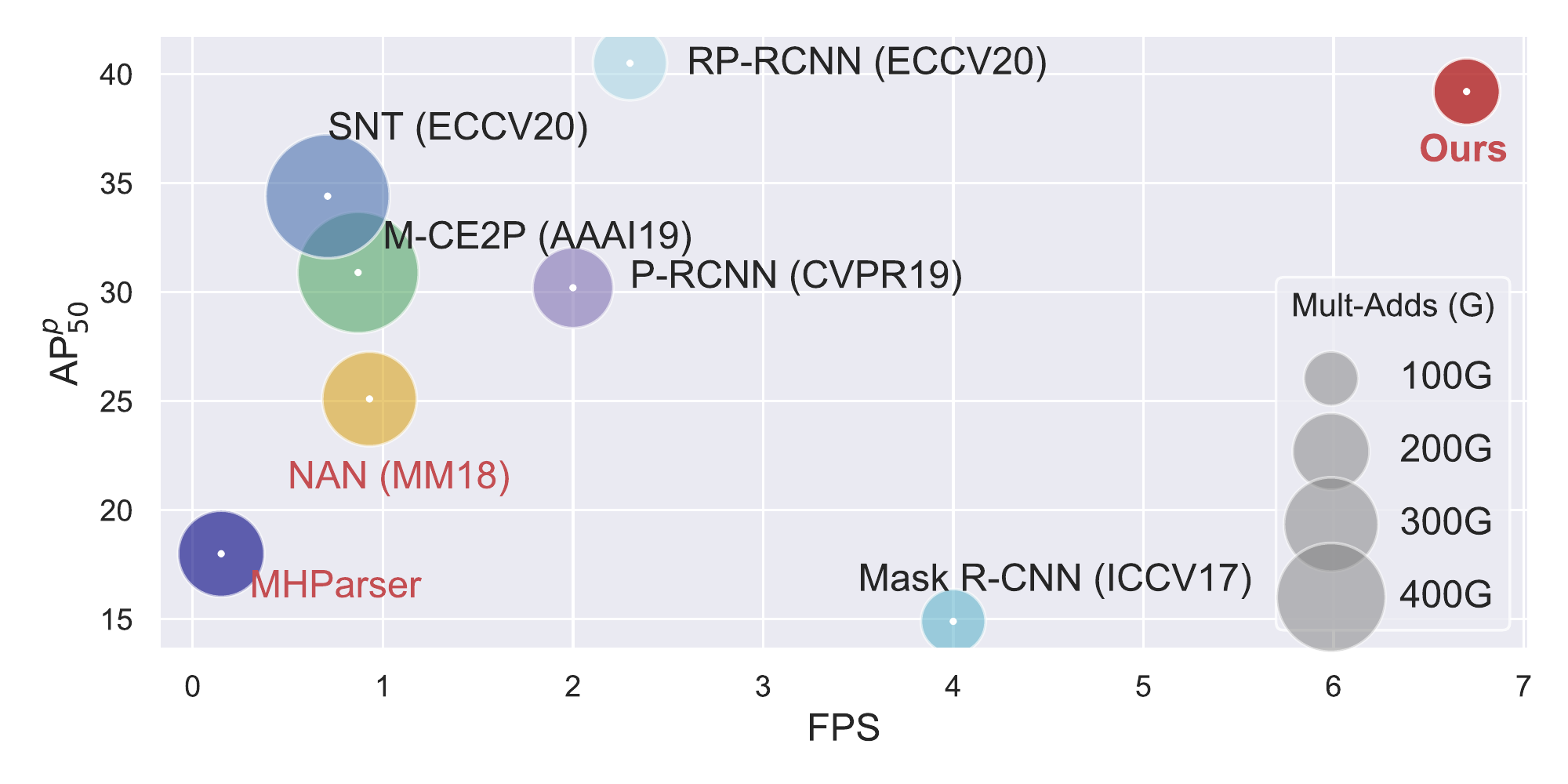}
	\end{center}
	\vspace{-18pt}
	\captionsetup{font=small}
	\caption{\!\small\textbf{Trade-off between performance \textit{vs.}\!~efficiency} on MHP$_{v2}$ \texttt{val}\!~\cite{zhao2018understanding}. The \textit{x}-axis and \textit{y}-axis denote FPS and AP$_{50}^p$, respectively. The circle size indicates Multi-Adds\!~(G). Top-down and bottom-up models are given in black and red, respectively. Our model shows promising performance with high efficiency.}
	\label{fig:madd}
	\vspace{-14pt}
\end{figure}

Concretely, three crucial techniques, for the first time in the field, are exploited to deliver our compact and powerful instance-aware human parsing solution:
\begin{itemize}[leftmargin=*]
	\setlength{\itemsep}{0pt}
	\setlength{\parsep}{-2pt}
	\setlength{\parskip}{-0pt}
	\setlength{\leftmargin}{-8pt}
	\vspace{-4pt}
	\item \textit{Differentiable body-to-joint projection}:\!~We propose a \underline{D}ense-to-\underline{S}parse \underline{P}rojection \underline{F}ield (DSPF), which is a set of 2D vectors over the image lattice. For each human pixel, its DSPF vector encodes the offset to the closest instance joint. DSPF thus allows us to explicitly associate dense human semantics with sparse joint representations.
	
	\item  \textit{Multi-step DSPF estimation}:\!~To address the difficulties of human body variations and occlusions, DSPF is computed in a coarse-to-fine fashion, over the network feature pyramid.  Deeper-layer features are low-resolution, yet robust to the challenges. They are thus used to derive an initial discriminative DSPF. Conditioning on the shallower-layer features, a finer DSPF is inferred by computing the residue to the coarse estimation. To reduce the feature-space distance across network layers, cross-layer feature alignment is learnt and adopted. In this way, the residue has a smaller magnitude and is easier to infer. 
	
	\item \textit{An end-to-end trainable framework}:\!~Current bottom-up multi-person pose estimators solve joint association through heuristic greedy algorithms~\cite{cao2017realtime,kreiss2019pifpaf}, independent from joint detection. This breaks the end-to-end pipeline and leads to suboptimal results.  Since joint association can be formulated as maximum-weight bipartite matching\!~\cite{cao2017realtime}, we explore a differentiable solution to this, inspired by\!~\cite{zeng2019dmm}. It revisits projected gradient descent and Dykstra's cyclic projection\!~\cite{boyle1986method} for convex constrained minimization. The$_{\!}$ solution$_{\!}$ is$_{\!}$ neat and light-$_{\!}$weight, allowing directly using the grouping errors for$_{\!}$ supervision.
	\vspace{-18pt}
\end{itemize}

Our$_{\!}$ model$_{\!}$ is$_{\!}$ also$_{\!}$ distinguished$_{\!}$ for$_{\!}$ its$_{\!}$ practical$_{\!}$ utility. First, it can benefit a host of human-centric applications. Some of them, such as augmented reality, are pose guided, while others, such as live streaming, video editing and virtual try-on systems, require understanding of fine-grained human semantics. Thus, our method can meet different needs in real-life applications with only a \textit{single} model. Second, due to its bottom-up nature and fast keypoint association, our method generates instance-level parsing results at a high speed of 0.15 s per image (see~\figref{fig:madd}), irrespective of the number of people in the scene, and, which is much faster than other alternatives (\eg, 1.15 s~\cite{ruan2019devil}, 14.94 s~\cite{li2017multiple}).

Extensive experiments are conducted on three instance-aware human parsing datasets. Specifically,~on MHP$_{v2}$\!~\cite{zhao2018understanding}, our model achieves an AP$_{50}^p$ of $39.0$, much$_{\!}$ better$_{\!}$ than$_{\!}$ current$_{\!}$ top-leading$_{\!}$ bottom-up method (\ie, NAN\!~\cite{zhao2018understanding} of $25.1$) and on par with best top-down parsers\!~\cite{ji2019learning,yang2020eccv}. On DensePose-COCO\!~\cite{alp2018densepose} and PASCAL-Person-Part\!~\cite{xia2017joint}, our model also produces compelling performance. Overall, our model shows favorable results with a fast speed. 

\vspace{-4pt}
\section{Related Work}
\vspace{-4pt}
\noindent\textbf{Instance-Aware Human Semantic Parsing}. Fine-grained human semantic segmentation, as one of the central tasks in human understanding, has applications in human-centric vision~\cite{qi2018learning,zhou2021cascaded,shen2019human}, human-robot interaction~\cite{fan2019understanding} and fashion analysis~\cite{wang2018attentive}. However, previous studies mainly focus on \textit{category-level} human parsing~\cite{liang2015human,gong2017look,fang2018weakly,liu2018cross,wang2019CNIF,zhang2020correlating,wang2020hierarchical}; only very few human parsers are specifically designed for the \textit{instance-aware} setting. 
As of to date, there exist two paradigms~for instance-aware human parsing: \textit{top-down} and \textit{bottom-up}. Top-down approaches\!~\cite{li2017holistic,yang2019parsing,ruan2019devil,ji2019learning,yang2020eccv} typically locate human instance proposals first, and then parse each proposal in a fine-grained manner. In contrast, bottom-up human parsers~\cite{gong2018instance,li2017multiple,zhao2018understanding} simultaneously perform pixel-wise instance-agnostic parsing and pixel grouping, inspired by existing bottom-up instance segmentation techniques. Their grouping strategies vary from instance-edge-aware clustering~\cite{gong2018instance}, to graph-based superpixel association~\cite{li2017multiple}, to proposal-free instance localization~\cite{zhao2018understanding}.

Our model falls into the bottom-up paradigm and has several unique characteristics. First, our method formulates instance-aware human part parsing as a multi-granularity human representation learning task. However, other alternatives, whether top-down or bottom-up, fail to explore skeleton-based human structures. Second, our model yields a new bottom-up regime for instance-aware human part parsing. Through projecting dense human semantics into sparse body joints, the task of pixel grouping can be easily tackled by joint association. Third, our method achieves promising results with much improved efficiency, leading to high utility in downstream applications.

%



\noindent\textbf{Multi-Person Pose Estimation.} Allocating body joints to human instances, as another representative human understanding task, has been extensively studied over the past decades. Also, current popular solutions for multi-person pose estimation can be roughly categorized as either \textit{top-down} or \textit{bottom-up} methods. Specifically, top-down methods\!~\cite{fang2017rmpe,huang2017coarse,papandreou2017towards,chen2018cascaded} conduct single-person pose estimation over each pre-detected human bounding-box. Contrarily, bottom-up methods are detection-free; they usually predict identity-free keypoints, which are then assembled into valid pose instances~\cite{insafutdinov2016deepercut,cao2017realtime,newell2017associative,kreiss2019pifpaf}, or directly regress poses from person instance centroids in a single-stage fashion~\cite{nie2019single}.

In our method, human joints are set as the targets for person pixel embedding learning, simplifying the procedure of pixel grouping. This flexible architecture can be seamlessly integrated with any bottom-up pose estimators in principle, and explicitly encodes human structural constraints. Hence, we formulate joint association as a \textit{differentiable} matching problem~\cite{zeng2019dmm}, rather than relying on sophisticated post-processing~\cite{cao2017realtime,kreiss2019pifpaf} like conventional bottom-up methods. Though\!~\cite{jin2020differentiable} also addresses joint association in an end-to-end manner, it needs to learn a complicated and heavy graph network and cannot guarantee optimality.



\noindent\textbf{Human/Object Representation.} From a slightly broader perspective, the way of representing visual elements is crucial for visual understanding. From classic rectangular boxes and masks, to recent point-based object modeling forms\!~(\eg, sparse points\!~\cite{yang2019reppoints}, contours\!~\cite{xie2020polarmask,wei2020point}, grid masks\!~\cite{chen2019tensormask}, and dense points\!~\cite{yang2019dense}), the community is continually pursuing a proper object representation for more effective processing, further advancing the development of object detection and segmentation. Point-based representations are thought to be promising, as both geometric and semantic cues are encoded. This insight is shared by several skeleton-based$_{\!}$ human$_{\!}$ understanding$_{\!}$ models. Despite the efforts made prior to the renaissance of deep learning~\cite{dong2014towards,xia2016pose}, some recent studies explore body poses as an extra representation granularity in category-level human parsing~\cite{xia2017joint,nie2018mutual} or instance-aware human full-body segmentation~\cite{papandreou2018personlab,zhang2019pose2seg}. However,~\cite{xia2017joint,nie2018mutual,zhang2019pose2seg} only utilize human poses as a shape prior for feature enhancement, instead of as an informative clue for person separation. Thus, their main ideas are far different from ours. Although\!~\cite{papandreou2018personlab} also views human joints as pixel embedding targets, our method \textbf{1)} focuses on fine-grained human semantic part parsing; \textbf{2)} coarse-to-fine estimates the projection from human dense semantics to sparse joints through residual learning and feature alignment; and \textbf{3)} formulates joint association in a differentiable manner, yielding an end-to-end trainable model.



\vspace{-2pt}
\section{Our Approach}
\vspace{-2pt}
As shown in~Fig.\!~\ref{fig:framework},~our model learns to: \textbf{1)} predict category-level human parts\!~(\S\ref{sec:iap}), \textbf{2)} build a Dense-to-Sparse Projection Field\!~(DSPF,\!~\S\ref{sec:dspf}), and \textbf{3)} conduct human joint detection and association\!~(\S\ref{sec:jda}). With~\textbf{1)}, we have fine-grained human semantics, but they are identification-irrelevant. With \textbf{2)}, we can explicitly associate dense human semantics with sparse joints, as DSPF encodes the displacements from each person pixel to the closet human joints. Instance-aware parsing can then be achieved by bottom-up joint association. At the same time, human structural cues can be embedded. Finally, with \textbf{3)}, joint association is achieved by a differentiable solver, allowing the error signals for grouping to be back-propagated to supervise keypoint detection and feature learning. In this way, we provide an instance-aware human parsing framework based on multi-granularity human representation learning, which works in a bottom-up manner and is  end-to-end trainable.

\vspace{-1pt}
\subsection{Instance-Agnostic Human Part Parsing}\label{sec:iap}
\vspace{-1pt}
Given an input image $I\!\!\in\!\!\mathbb{R}^{W\!\times\!H\!\times\!3}$, a backbone network is first employed to extract an $L$-level feature pyramid, \ie, $\{\bm{X}_l\!\in\!\mathbb{R}^{W_l\!\times\!H_l\!\times\!C_l}\}_{l=1}^L$, where $W_l=W/2^{l+1}$ and $H_l=H/2^{l+1}$. The feature pyramid
comprehensively encodes multi-scale visual features, from the highest spatial resolution ($l\!=\!1$) to the lowest ($l\!=\!L$).

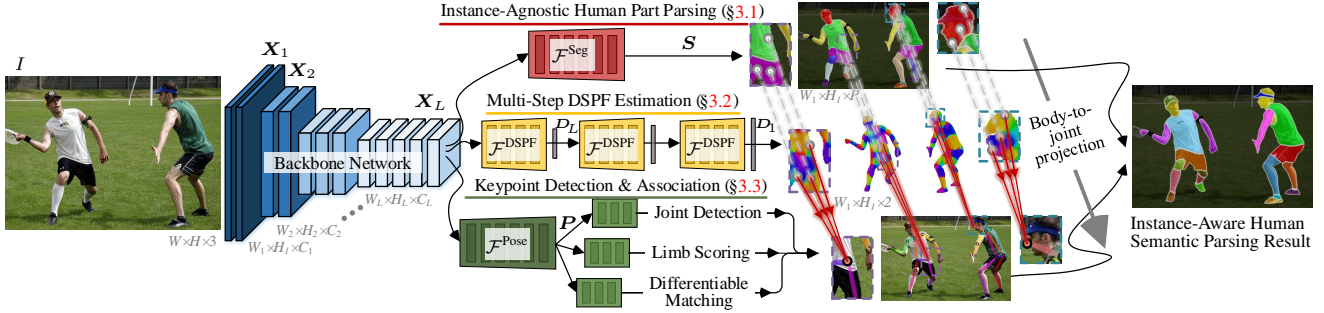
\begin{figure*}[t]
	\vspace{-5pt}
	\begin{center}
		\includegraphics[width=\linewidth]{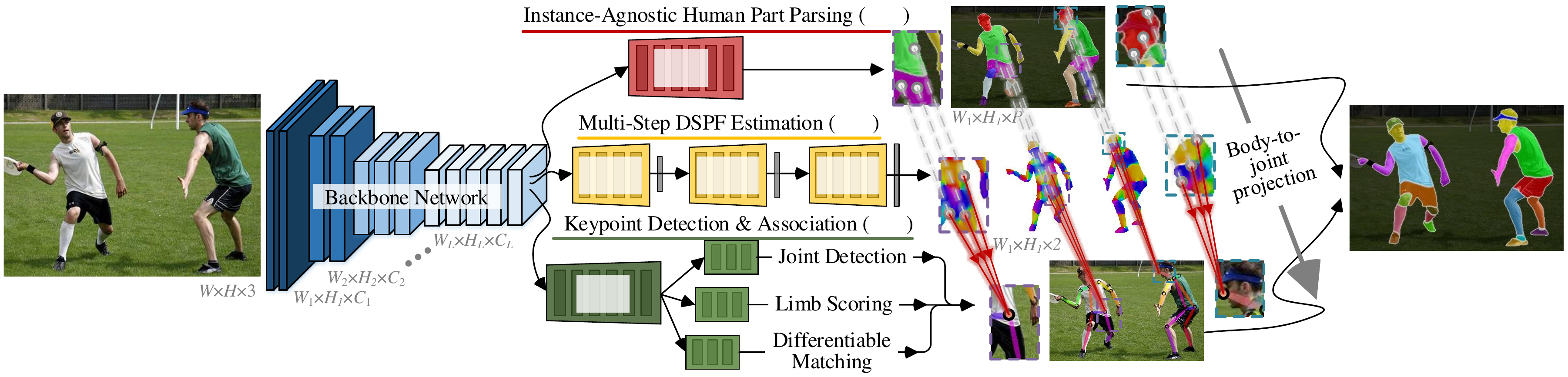}
		\put(-492,92){{\scriptsize$I$}}
		\put(-223,112){{\scriptsize\S\ref{sec:iap}}}
		\put(-289,95){\scriptsize$\mathcal{F}^{\text{Seg}}$}
		\put(-233,78){\scriptsize\S\ref{sec:dspf}}
		\put(-313,60){\scriptsize$\mathcal{F}^{\text{DSPF}}$}
		\put(-276,60){\scriptsize$\mathcal{F}^{\text{DSPF}}$}
		\put(-238,60){\scriptsize$\mathcal{F}^{\text{DSPF}}$}
		\put(-221.9,46){\scriptsize\S\ref{sec:jda}}
		\put(-314,23){\scriptsize$\mathcal{F}^{\text{Pose}}$}
        \put(-399,98){\scriptsize$\bm{X}_1$}
        \put(-389,90){\scriptsize$\bm{X}_2$}
        \put(-341,78){\scriptsize$\bm{X}_L$}
        \put(-286.4,32){\scriptsize$\bm{P}$}
        \put(-240,100){\scriptsize$\bm{S}$}
        \put(-288,70){\tiny$\bm{D}_{\!L}$}
        \put(-212,70){\tiny$\bm{D}_{\!1}$}
        \put(-70,32){\scriptsize Instance-Aware Human}
        \put(-70,25){\scriptsize Semantic Parsing Result}
	\end{center}
	\vspace{-15pt}
	\captionsetup{font=small}
	\caption{\small\textbf{Illustration of our multi-granularity human representation learning framework} for instance-aware human semantic parsing.}
	\label{fig:framework}
	\vspace{-12pt}
\end{figure*}

Given $\{\bm{X}_l\}_{l=1}^L$, a segmentation head $\mathcal{F}^{\text{Seg}}$ is applied to parse category-level human part semantics (see Fig.~\ref{fig:framework}):
\begin{equation}\small
\begin{aligned}\label{eq:1}
\bm{S} = \mathcal{F}^{\text{Seg}}(\{\bm{X}_l\}_{l=1}^L)\!\in\![0,1]^{W_1\!\times\!H_1\!\times\!P}.
\end{aligned}
\end{equation}
Here, $P$ indicates the number of human semantic part categories. $\mathcal{F}^{\text{Seg}\!}$ is implemented as a decoder architecture\!~\cite{chen2018encoder}, which makes full use of the multi-scale features $\{\bm{X}_l\}_{l=1}^L$ and estimates $\bm{S}$ with the highest spatial resolution.

\vspace{-1pt}
\subsection{Multi-Step Dense-to-Sparse Projection Field (DSPF) Estimation}\label{sec:dspf}
\vspace{-0pt}
Given $\{\bm{X}_l\}_{l=1}^L$, our network learns DSPF, \ie, $\bm{D}\!\in\!\mathbb{R}^{W_1\!\times\!H_1\!\times\!2}$, to associate dense human semantics with sparse keypoints\!~(Fig.\!~\ref{fig:framework}). Specifically, for each point $\bm{u}\!=\!(u, v)$ in an image lattice $\Omega\!\in\!\mathbb{R}^{2}$, its DSPF is a 2D vector $\bm{D}(\bm{u})\!=\!(\Delta u, \Delta v)$, where $\bm{u}\!+\!\bm{D}(\bm{u})\!=\!(u\!+\!\Delta u, v\!+\!\Delta v)$ is expected to be the location of its nearest human instance joint. Note that $\bm{u}$ and $\bm{u}\!+\!\bm{D}(\bm{u})$ belong to the same human instance. Due to variations and deformations typically exhibited by human bodies, as well as ambiguities caused by close or occluded instances, it is difficult to directly infer DSPF. Thus, we estimate it in a coarse-to-fine framework.

\noindent\textbf{Coarse-to-Fine Estimation.} Given the coarsest feature (\ie, $\bm{X}_L$), which has the largest receptive field and thus is good at robust, long-range relation modeling, an initial DSPF estimation $\bm{D}_L\!\in\!\mathbb{R}^{W_L\!\times\!H_L\!\times\!2}$ can be derived. Starting with $\bm{D}_L$, the most straightforward strategy is to incrementally update $\bm{D}_L$ across the pyramid levels to finally achieve a high-resolution DSPF estimation $\bm{D}_{1\!}\!\in\!\mathbb{R}^{W_{1}\!\times\!H_{1}\!\times\!2}$:
\vspace{-3pt}
\begin{equation}\small
\begin{aligned}\label{eq:dspf}
\bm{D}_l = \mathcal{F}^{\text{DSPF}}(\bm{X}_l, {\bm{X}}^{\uparrow}_{l+1}, 2\bm{D}_{l+1}^{\uparrow})\!\in\!\mathbb{R}^{W_l\!\times\!H_l\!\times\!2},
\end{aligned}
\vspace{-2pt}
\end{equation}
where $\bm{D}_{l+1\!}$ and ${\bm{X}}_{l+1\!}$ from a preceding level need to be upsampled
in spatial resolution\!~(denoted by ``$\uparrow$''), and $\bm{D}_{l+1}$ needs to be multiplied by 2 to match the resolution of the pyramidal features in the $l$-th level. In this way, the estimation for DSPF can be gradually improved by fusing higher-level prediction $\bm{D}_{l+1}$ with lower-layer feature $\bm{X}_l$. Since the flow magnitude is enlarged after each estimation step, the value ranges of output spaces are different, making the regression difficult. Inspired by recent optical flow estimation techniques~\cite{hui2018liteflownet,hur2019iterative}, we employ a \textit{residual refinement} strategy for more accurate DSPF estimation.

\noindent\textbf{Coarse-to-Fine Residual Refinement.} Rather than inferring a complete DSPF $\bm{D}_{l\!}$ at each pyramid level, our network learns the residue $\triangle\bm{D}_{l\!}\!\in\!\mathbb{R}^{W_l\!\times\!H_l\!\times\!2\!}$ in relative to $\bm{D}_{l+1\!}$ estimated in the previous layer. Eq.\!~\eqref{eq:dspf} is thus improved as:
\vspace{-3pt}
\begin{equation}\small
\begin{aligned}\label{eq:res}
\bm{D}_l = \underbrace{\mathcal{F}^{\text{DSPF}}(\bm{X}_l, {\bm{X}}^{\uparrow}_{l+1})}_{\triangle\bm{D}_{l}}+2\bm{D}_{l+1}^{\uparrow}.
\end{aligned}
\vspace{-5pt}
\end{equation}
This network design is more favored, since the magnitudes of the residues at different pyramid levels are smaller than the complete DSPFs and within a similar value range.

Recall that we apply ``$\uparrow$'' operation to warp a low-resolution feature map into a high-resolution one. A common way to implement such a warping operation is through bilinear upsampling. However, as bilinear upsampling only accounts for limited information from a fixed and pre-defined subpixel neighborhood for feature interpolation, it fails to handle the misalignment between feature maps caused by repeated padding, downsampling and upsampling, thus leading to error accumulation. Motivated by learnable feature warping operators\!~\cite{sun2018pwc,wang2019carafe}, we propose to learn a feature flow between two features from adjacent pyramid levels for cross-layer feature alignment.

\noindent\textbf{Learnable Feature Warping.}  In particular, a feature flow, \ie, $\bm{F}_{l}$, is estimated between $\bm{X}_{l}$ and $\bm{X}_{l+1}$:
\vspace*{-4pt}
\begin{equation}\small
\begin{aligned}\label{eq:featureflow}
 \text{feature flow:}&~~~\bm{F}_{l} = \mathcal{F}^{\text{Flow}}(\bm{X}_{l}, \bm{X}_{l+1}^{~\!\!\Uparrow}) \in \mathbb{R}^{W_l\!\times\!H_l\!\times\!2}.
\end{aligned}
\vspace*{-4pt}
\end{equation}
Here, the low-resolution feature $\bm{X}_{l+1}$ is first bilinearly upsampled (denoted by ``$\Uparrow$''), and then concatenated with the high-resolution feature $\bm{X}_{l}$ for predicting $\bm{F}_{l}$. The feature flow, in essence, is an offset field that aligns the positions between high-level and low-level feature maps. Through $\bm{F}_{l}$,  $\bm{X}_{l+1\!}$ is warped towards $\bm{X}_{l}$, \ie, $\bm{X}_{l+1\!}^{~\!\!\uparrow}(\bm{u}\!+\!\bm{F}_{l}(\bm{u}))\!\sim\!\bm{X}_{l}$. Hence, we have:
\vspace*{-4pt}
\begin{equation}\small
\begin{aligned}
\text{feature warping:}&~~~{\bm{X}}^{\uparrow}_{l+1} = \mathcal{F}^{\text{Warp}}({\bm{X}}_{l+1}, \bm{F}_{l}) \in \mathbb{R}^{W_l\!\times\!H_l\!\times\!C_l}, \\
\text{DSPF warping:}&~~~\bm{D}_{l+1}^{\uparrow} = \mathcal{F}^{\text{Warp}}(\bm{D}_{l+1}, \bm{F}_{l}) \in \mathbb{R}^{W_l\!\times\!H_l\!\times\!2}.
\end{aligned}\label{eq:fdwarp}
\vspace*{-4pt}
\end{equation}
The warping operation $\mathcal{F}^{\text{Warp}\!}$ upsamples the input (${\bm{X}}_{l+1\!}$  or $\bm{D}_{l+1}$) at the $(l\!+\!1)$-th level to match the high resolution of the $l$-th level, according to the estimated feature flow $\bm{F}_{l}$. Taking the feature warping in Eq.\!~\ref{eq:fdwarp} as an example, we have:
\vspace*{-4pt}
\begin{equation}\small
\begin{aligned}\label{eq:warp}
{\bm{X}}^{\uparrow}_{l+1}(\bm{u})\!=\!\!\!\!\sum_{\bm{u}'_s\in\mathcal{N}(\bm{u}_s)}\!\!\!\!{\bm{X}}^{~\!\!\Uparrow}_{l+1}(\bm{u}'_s)(1\!-\!|u_s \!-\! u'_s|)(1\!-\!|v_s\!-\!v'_s|),
\end{aligned}
\vspace*{-2pt}
\end{equation}
where $\bm{u}_s\!=\!\bm{u}\!+\!\bm{F}_{l}(\bm{u})\!=\!(u_s, v_s)$ denotes the source coordinate in ${\bm{X}}^{~\!\!\Uparrow}_{l+1}$, $\bm{u}$ indicates the target coordinate in the interpolated output (\ie, ${\bm{X}}^{\uparrow}_{l+1}$), and $\mathcal{N}(\bm{u}_s)$ is the four pixel neighbors of $\bm{u}_s$. Note that $\mathcal{F}^{\text{Warp}}$ is differentiable, and its gradient can be effectively computed following~\cite{jaderberg2015spatial}.

\subsection{Fully Differentiable Human Keypoint Detection and Association}\label{sec:jda}

So far, we have obtained the instance-agnostic human semantic part estimation $\bm{S}$\!~(\S\ref{sec:iap}) and DSPF\!~$\bm{D}$ (\S\ref{sec:dspf}). Now we aim to precisely detect human joints and assemble them into human instances. By doing this, we can infer the instance-aware parsing result from $\bm{S}$, according to $\bm{D}$\!~(\ie, the projection from dense human semantics to sparse keypoints; Fig.\!~\ref{fig:framework}). Hence, by formulating joint association as maximum-weight bipartite matching, a differentiable solution is derived to make our model end-to-end trainable.

Prior to pose estimation, we apply a small decoder over $\{\bm{X}_l\}_{l=1}^L$ to learn a pose-specific feature representation $\bm{P}$:
    \vspace{-4pt}
\begin{equation}\small
\begin{aligned}\label{eq:pose}
~~~~~~~~\bm{P} = \mathcal{F}^{\text{Pose}}(\{\bm{X}_l\}_{l=1}^L\}) \in \mathbb{R}^{W_1\!\times\!H_1\!\times\!C_1}.
\end{aligned}
    \vspace{-4pt}
\end{equation}
$\bm{P}$ merges multi-scale cues and maintains a high resolution.

\begin{figure}[t]
	\begin{center}
		\includegraphics[width=\linewidth]{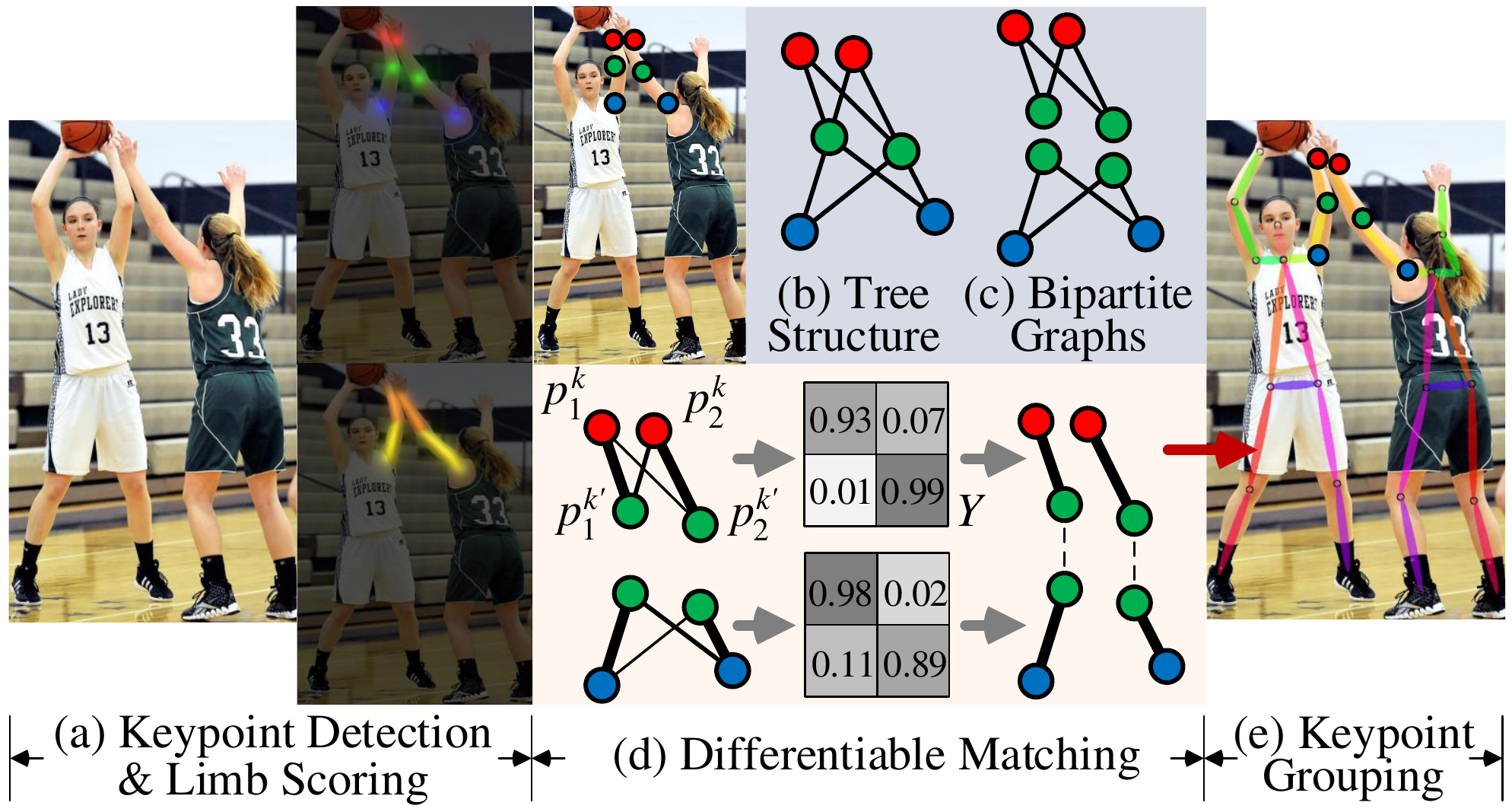}
	\end{center}
	\vspace{-15pt}
	\captionsetup{font=small}
	\caption{\small\textbf{Illustration of our differentiable solver for human keypoint detection and association.} See \S\ref{sec:jda} for details.}
	\label{fig:pose}
	\vspace{-15pt}
\end{figure}

\noindent\textbf{Keypoint Detection.}
We first predict the locations of all the visible anatomical keypoints (\eg, left shoulder, right elbow) in $I$.
Following~\cite{papandreou2017towards,kreiss2019pifpaf}, we achieve this by jointly predicting confidence maps and 2D local offset fields over $\bm{P}$. Assume there are a total of $K$ keypoint categories, and, for each category $k$, we compute a heatmap $M\!\in\![0,1]^{W_1\!\times\!H_1}$, where ${M}(\bm{u})\!=\!1$  if the location $\bm{u}\!=\!(u,v)$ falls into a disk of radius $R$ ($=\!32$ pixels) of any keypoint of category $k$; otherwise $M(\bm{u})\!=\!0$. Moreover, for each category $k$, we compute an offset field $\bm{O}\!\in\!\mathbb{R}^{W_1\!\times\!H_1\!\times\!2}$ to improve the localization, where each offset $\bm{O}(\bm{u})$ points from $\bm{u}$ to its closest keypoint of category $k$. The heatmap $M$ and offset field $\bm{O}$ are aggregated by Hough voting to obtain a highly localized Hough score map $H\!\in\!{[0, 1]}^{W_1\!\times\!H_1}$, whose element at location $\bm{u}$ is computed as:
    \vspace{-5pt}
\begin{equation}\small
\begin{aligned}
\!\!H(\bm{u})\! = \!\sum\nolimits_{\bm{u}'\in\Omega} \frac{1}{\pi\!R^2} M(\bm{u})\mathcal{F}^{\text{Bi-inter}}(\bm{u}' + \bm{O}_k(\bm{u}') - \bm{u}),
\end{aligned}
    \vspace{-3pt}
\end{equation}
where $\mathcal{F}^{\text{Bi-inter}}$ is a bilinear interpolation function. The local maxima in $H$ correspond to the keypoints of the $k$-th category. We compute $K$ heatmaps and offset fields and conduct Hough voting individually for each category (see Fig.~\ref{fig:pose}\!~(a)). Through DSPF, we can get a set of keypoint coordinates, \ie, $\{\bm{u}\!+\!\bm{D}(\bm{u})\}_{\bm{u}\in\Omega}$. With localized Hough score map of each category, we can determine the score at each of these coordinates. Finally, the coordinates whose scores are larger than $0.7$ are preserved as the detected keypoints.

\noindent\textbf{Limb Scoring.} We denote $\mathcal{P}\!=\!\{\bm{p}^k_n\!: k\!\in\{1,\ldots,K\},\!~n\!\in\!\{1,\ldots,N_k\}\}$ as the set of keypoints detected in $I$, where $N_k$ is the number of keypoints belonging to the $k$-th category, and  $\bm{p}^k_n\!=\!(u^k_n,v^k_n)$ represents the 2D coordinate of the $n$-th detected keypoint of the $k$-th category. The detected joints $\mathcal{P}$ serve as candidate positions for human poses, and provide us possible kinematic connections (\ie, limbs) so that we can assemble $\mathcal{P}$ to form full body poses. To do so, we first need to score each possible limb hypothesis. For each limb category, we predict a limb heatmap $Q\!\in\!\mathbb{R}^{W_1\!\times\!H_1}$ over $\bm{P}$, which represents the confidence of the limb\!~(see Fig.~\ref{fig:pose}\!~(a)). For simplicity, we approximately represent each limb using an elliptical area between the endpoints\!~\cite{li2020simple}, and generate the ground-truths by applying an unnormalized elliptical Gaussian distribution with a standard deviation $\sigma$\!~($=\!9$ pixels) over all limbs. The score of each limb is then estimated by sampling a set of Gaussian responses within the limb area, from the corresponding heatmap $Q$. Thus, for each limb with two joint categories $k$ and $k'$, we can get a scoring matrix $A\!\in\!{[0, 1]}^{N_k\!\times\!N_{k'}}$, whose element $a_{nn'}$
stores the connectivity between joints $\bm{p}^k_n$ and $\bm{p}^{k'}_{n'}$.

\noindent\textbf{Differentiable Matching.}
The keypoint candidates $\mathcal{P}$ define a large set of possible
limbs. We score each limb to determine the likelihood of it belonging to a particular person, according to which we parse optimal human poses (\ie, best joint matching).
As shown in Fig.~\ref{fig:pose}\!~(b), the problem of finding the optimal matching corresponds to a $K$-dimensional matching problem that is known to be NP-hard~\cite{west1996introduction}. Thus, we decompose the matching problem into a set of bipartite matching subproblems~\cite{cao2017realtime}, \ie, determine the matching for each connected limb independently (see Fig.~\ref{fig:pose}\!~(c)). Specifically, in each bipartite graph, a matching corresponds to a subset of edges in which no two edges share a node.
Formally, we denote $\mathcal{P}^{k\!}\!=\!\{\bm{p}^{k}_n\}_{n=1}^{N_k}$ and $\mathcal{P}^{k'\!}\!=\!\{\bm{p}^{k'}_{n'}\}_{n'=1}^{N_{k'}}$ as two sets of detected keypoints belonging to category $k$ and $k'$, respectively. We further define a variable $y_{nn'}\!\in\!\{0,1\}$ to represent whether $\bm{p}^{k}_n$ and $\bm{p}^{k'}_{n'}$ are connected as a limb, and our goal becomes finding the optimal assignment  $\{y_{nn'\!}\}_{n,n'}$ for the set of all possible connections. For each limb with joint endpoint categories $k$ and $k'$ and the corresponding scoring matrix $A\!\in\!{[0, 1]}^{N_k\!\times\!N_{k'}}$, we infer the  boolean assignment matrix ${Y}\!\in\!\{0,1\}^{N_k\!\times\!N_{k'}}$ by solving the maximum-weight bipartite matching problem:
\par\nobreak
\vspace{-14pt}
{\small\setlength{\belowdisplayskip}{4pt}
\begin{align}
	& \max_{{Y}} \sum\nolimits_{n=1}^{N_k} \!\sum\nolimits_{n'=1}^{N_{k'}} \!a_{nn'}\cdot y_{nn'}, \label{eq:constraint0}\\
	\textit{s.t.}~~~~ & \forall{n},~~~~\sum\nolimits_{n'=1}^{N_{k'}}y_{nn'} \leq 1, \label{eq:constraint1}\\
	& \forall{{n'}},~~~~\sum\nolimits_{n=1}^{N_k}y_{nn'} \leq 1,
	\label{eq:constraint2} \\
	&\forall(n, {n'}), ~~~~y_{nn'\!}\in\{0,1\},\label{eq:constraint3}
\end{align}}%
where $a_{nn'}\!\in\!\bm{A}$ is the connection score between joints $\bm{p}^{k}_{n}$ and $\bm{p}^{k'}_{n'}$. The constraints in Eqs.~(\ref{eq:constraint1}-\ref{eq:constraint2}) ensure that no two edges will share a node. The final multi-person pose parsing result is with maximum weight for the chosen edges (\ie, connection scores). This integer linear programming problem can be solved using Hungarian algorithm~\cite{kuhn1955hungarian} with a high time complexity of $O(n^3)$. Existing pose estimators~\cite{cao2017realtime,papandreou2018personlab,kreiss2019pifpaf} instead employ heuristic greedy algorithms, but break the end-to-end pipeline. Inspired by~\cite{zeng2019dmm}, we propose a differentiable solution which facilitates model learning with direct matching based supervision.

In particular, we drop the integer constraint (Eq.\!~\eqref{eq:constraint3}) and compute $Y$ as a real-valued assignment matrix, by solving a linear programming relaxation and rewriting Eqs.~(\ref{eq:constraint0}-\ref{eq:constraint3}) in matrix form:
\vspace{-22pt}
\par\nobreak
{\small\setlength{\belowdisplayskip}{6pt}
\begin{align}
	& \min_{{Y}} \text{Tr}(-{A}{Y}^\top), \label{eq:tf1} \\
\!\!\!\!\!\!\!\!\!\textit{s.t.}~~~~ {Y}\textbf{1}_{N_{k'}} \leq \textbf{1}_{N_k}, &~~~~{Y}^\top\textbf{1}_{N_{k}} \leq \textbf{1}_{N_{k'}}, ~~~~{Y} \geq 0.\label{eq:tf2}
	\end{align}
}%
Here, $\textbf{1}_{N_{k}\!}\!=\![1]^{N_{k}\times1\!}$ and $\textbf{1}_{N_{k'}\!}\!=\![1]^{N_{k'}\times1\!}$. Since both the target function (Eq.\!~\eqref{eq:tf1}) and constraint functions (Eq.\!~\eqref{eq:tf2}) are convex, we can solve this convex constrained minimization problem using projected gradient descent (PGD)~\cite{calamai1987projected}. Let us denote the constraints in Eq.\!~\eqref{eq:tf2} as $\mathcal{C}\!=\!\mathcal{C}_1\!\cap\!\mathcal{C}_2\!\cap\!\mathcal{C}_3$, where $\mathcal{C}_1\!=\!{Y}\textbf{1}_{N_{k'\!}}\!\leq\!\textbf{1}_{N_k}$, $\mathcal{C}_2\!=\!{Y}^{\top\!}\textbf{1}_{N_{k}\!}\!\leq\!\textbf{1}_{N_{k'}}$, and $\mathcal{C}_3\!=\!{Y}\!\geq\!0$. PGD estimates $Y$ by iterating the following equation:
\vspace{-3pt}
\begin{equation}\small
\begin{aligned}
Y\leftarrow\mathcal{F}^{\mathcal{C}\!}(Y-\alpha\triangledown f(Y)),
\end{aligned}
\vspace{-3pt}\label{eq:pgd}
\end{equation}
where $f(Y)\!=\!\text{Tr}(-AY^{\top\!})$, $\triangledown f(Y)\!=\!-A$, and $\alpha$ indicates the step size. Here $\mathcal{F}^{\mathcal{C}\!}$ is a projection operator, and itself is also an optimization problem:
\vspace{-5pt}
\begin{equation}\small
\begin{aligned} \mathcal{F}^{\mathcal{C}\!}(Y)\!=\!\argmin_{Y'\in\mathcal{C}}\frac{1}{2}||Y'\!-\!Y||_2^2.
\end{aligned}
\vspace{-5pt}
\end{equation}
Given $Y$, $\mathcal{F}^{\mathcal{C}}$ tries to find a ``point'' $Y'\!\in\!\mathcal{C}$ which is closet to $Y$. The challenge here is how to find a feasible projection operator over the combination of several convex constraints, \ie, $\mathcal{C}_{1}\cap\mathcal{C}_{2}\cap\mathcal{C}_3$. A popular method for finding the projection onto the intersection of convex subsets in Hilbert space is the Dykstra's cyclic projection algorithm~\cite{dykstra1983algorithm,boyle1986method}. It reduces the problem to an iterative scheme which involves only finding projections from the individual sets. As the projection operator with respect to each constraint can be easily derived in a differentiable form, Eqs.~(\ref{eq:tf1}-\ref{eq:tf2}) can be efficiently solved by PGD. In this way, our whole human part parsing model is end-to-end trainable and can direct access to keypoint matching supervision signals. After obtaining the assignment matrices for all the limbs, we follow conventions~\cite{cao2017realtime,kreiss2019pifpaf} to sequentially group the matched keypoints from ``head'' to ``foot'', as the estimations of ``head'' are usually more accurate than other keypoints. Finally, multi-person pose parsing results can be delivered (see Fig.~\ref{fig:pose}\!~(d-e)).

\vspace{-2pt}
\subsection{Implementation Details}\label{sec:nd}
\vspace{-1pt}
\noindent\textbf{Loss Function.} For category-level human semantic prediction $\bm{S}$\!~(\S\ref{sec:iap}), \textit{cross-entropy} loss is used.  For DSPF estimation\!~(\S\ref{sec:dspf}), $l_1$ loss is used.  For keypoint detection and joint affinity estimation\!~(\S\ref{sec:jda}), \textit{cross-entropy} loss is used for confidence/score map and assignment matrix prediction, and $l_1$ loss is used for vector regression. The weights of different losses are adaptively adjusted during training by considering the homoscedastic uncertainty of each task~\cite{kendall2018multi}.

\noindent\textbf{Network Configuration.}
We compute the feature pyramid$_{\!}$ $\{_{\!}\bm{X}_{l\!}\}_{l=1\!}^L$ at$_{\!}$ $L\!\!=\!\!4$ levels$_{\!}$ with$_{\!}$ a$_{\!}$ scaling$_{\!}$ step$_{\!}$ of$_{\!}$ $2$~\cite{lin2017feature}. We adopt ResNet-101~\cite{he2016deep} as the backbone. The segmentation head $\mathcal{F}^{\text{Seg}\!}$ in Eq.~(\ref{eq:1}) has a similar architecture as the decoder in~\cite{chen2018encoder}, while in each upsampling stage we apply a single $5\!\times\!5$ depthwise-separable convolution to build a lightweight module. The pose head $\mathcal{F}^{\text{Pose}\!}$ in Eq.~\eqref{eq:pose} is built in a similar way. For $\mathcal{F}^{\text{Flow}\!}$ in Eq.\!~(\ref{eq:featureflow}), it concatenates the inputs and then applies a $3\!\times\!3$ convolutional layer for feature flow prediction.

\noindent\textbf{Training.}
We train our model in two steps. In the first step, the model is trained without differentiable matching using the SGD optimizer with a mini-batch size of $32$. We start the training with a base learning rate of 5e-2, momentum of $0.9$ and weight decay of 1e-5. The learning rate is then reduced to 5e-3 after $120$ epochs, and the training is terminated within $150$ epochs. In the second step, we finetune the whole model for another $50$ epochs with a fixed learning rate of 1e-4. During all training stages, we fix the running statistics of batch normalization layers to the ImageNet-pretrained values. Typical data augmentation techniques, \eg, horizontal flipping, scaling in range of [0.5, 2], rotation from [-10\si{\degree}, 10\si{\degree}], and color jitter, are also used.

\noindent\textbf{Inference.}
Once trained, our model can be directly applied to unseen images. For a test image, we detect a set of key-points $\mathcal{P}\!=\!\{\bm{p}^k_n\}$ and group them into different human instances. For each human instance $h$, we denote the set of its keypoints as $\mathcal{P}_h\!=\!\{\bm{p}_{n,h}\}_n$. Then we associate each person pixel (identified by the category-level semantic parsing module) to one of the human instances. Formally, for each human pixel $\bm{u}$ and its DSPF vector $\bm{D}(\bm{u})$, an assignment score for each human instance $h$ is computed as:
\vspace{-2pt}
\begin{equation}\small
\psi_h = \min_{\bm{p}_{n,h}\in\mathcal{P}_h} (\frac{\|\bm{u}+ \bm{D}(\bm{u}) - \bm{p}_{n,h}\|_2}{(s_{n,h} + s_h) * \sigma_h}).
\vspace{-2pt}
\end{equation}
For each keypoint $\bm{p}_{n,h\!}$ belonging to the human instance~$h$, its distance to $\bm{u}\!+\!\bm{D}(\bm{u})$, \ie, $\bm{u}$'s associated human instance keypoint, is first computed and normalized by joint score$_{\!}$ (\ie,$_{\!}$ $s_{n,h\!\!}\!=_{\!}\!\!H(\bm{p}_{n,h})$), instance$_{\!}$ score (\ie,$_{\!}$ $s_{h\!}\!=_{\!}\!\frac{1}{|\mathcal{P}_h|}$ $\sum_{\bm{p}_{n,h}\in\mathcal{P}_h\!}H(\bm{p}_{n,h})$), and instance scale (\ie, $\sigma_h$) for reliable, scale-invariant evaluation. $\sigma_h$ is set as the square root of the area of the bounding box tightly containing $\mathcal{P}_h$. Finally, the human instance for $\bm{u}$ is assigned as: $\arg\min_h \psi_h$.

\vspace{-2pt}
\section{Experiment}
\vspace{-2pt}
\subsection{Experimental Setup}
\vspace{-2pt}
\noindent\textbf{Datasets:}~Our experiments are conducted on three datasets:\!\!
\begin{itemize}[leftmargin=*]
	\setlength{\itemsep}{0pt}
	\setlength{\parsep}{-2pt}
	\setlength{\parskip}{-0pt}
	\setlength{\leftmargin}{-10pt}
	\vspace{-6pt}
	\item \textbf{MHP$_{v2}$}\!~\cite{zhao2018understanding}
	is currently the largest dataset for instance-aware human parsing. In total,  it includes $25,\!403$ images ($15,\!403/5,\!000/5,\!000$ for \texttt{train}/\texttt{val}/\texttt{test} splits). Each person is elaborately annotated with $58$ fine-grained semantic categories (\ie, $11$ body parts, $47$ cloth and accessory labels), as well as $16$ body joints.
	
	\item \textbf{DensePose-COCO}\!~\cite{alp2018densepose} has $27,\!659$ images ($26,\!151$/$1,\!508$ for \texttt{train}/\texttt{test} splits) gathered from COCO\!~\cite{lin2014microsoft}. It annotates $14$ human parts and $17$ keypoints.
	
	\item \textbf{PASCAL-Person-Part}\!~\cite{xia2017joint} has $1,\!716$ and $1,\!817$ images for \texttt{train} and \texttt{test}, respectively, with annotations of $6$ human part categories and $14$ body joints.
	
	\vspace{-4pt}
\end{itemize}
\noindent\textbf{Evaluation Metrics.}$_{\!}$ For$_{\!}$ instance-agnostic$_{\!}$ parsing, we$_{\!}$ adopt mean intersection-over-union (mIoU). For instance-level parsing, we employ official metrics of each dataset for fair comparison. Specifically, for MHP$_{v2}$ and DensePose-COCO, the average precision based on parts (AP$^p$) and percentage of correctly parsed semantic parts (PCP) are used. For PASCAL-Person-Part, we follow conventions\!~\cite{hariharan2014simultaneous,xia2016zoom} to report the average precision based on regions (AP$^r$).


\begin{table}
	\centering
	\small
	\resizebox{0.49\textwidth}{!}{
		\setlength\tabcolsep{6pt}
		\renewcommand\arraystretch{1.05}
		\begin{tabular}{r||c|ccc|c}\thickhline
			\rowcolor{mygray}
			Methods & mIoU & AP$_{50}^p$ & AP$_{vol}^p$ & PCP$_{50}$  &  M-Adds\!~(G) \\ \hline\hline
			\multicolumn{6}{l}{\textit{top-down models:}} \\ \hline
			M-RCNN\!~\cite{he2017mask} & - & 14.9   & 33.9 & 25.1  &  141.3\\
			P-RCNN\!~\cite{yang2019parsing} & 40.3 & 30.2 & 41.8   & 44.2  & 220.7 \\
			M-CE2P\!~\cite{ruan2019devil} & 41.1 & 30.9 & 41.3   & 40.6  &  497.1 \\
			SNT\!~\cite{ji2019learning} & - & 34.4 & 42.5  & 43.5  & 520.5 \\
			RP-RCNN\!~\cite{yang2020eccv}  & 37.3 & 40.5 & 45.2  & 39.2  & 185.5 \\ \hline\hline
			\multicolumn{6}{l}{\textit{bottom-up models:}} \\ \hline
			PGN\!~\cite{gong2018instance} & 25.3 & 17.6 & 35.5 & 26.9  & 169.2 \\
			MHParser\!~\cite{li2017multiple} & - & 18.0 & 36.1   & 27.0   &- \\
			NAN\!~\cite{zhao2018understanding} & - & 25.1 & 41.8   & 32.3   & 302.2 \\
			\textbf{Ours} & 41.4 & 39.0 & 44.3  & 42.3  &151.1\\\hline
		\end{tabular}
	}
	\captionsetup{font=small}
	\caption{\small \textbf{Quantitative performance comparison on MHP}$_{v2}$ \texttt{val}\!~\cite{zhao2018understanding}, with mIOU, AP$^p$ and PCP. See \S\ref{sec:qr} for details.}
	\label{table:mhp}
	\vspace{-6pt}
\end{table}
\begin{table}
	\centering
	\small
	\resizebox{0.45\textwidth}{!}{
		\setlength\tabcolsep{10pt}
		\renewcommand\arraystretch{1.05}
		\begin{tabular}{r||c|ccc}
			\rowcolor{mygray}
			\thickhline
			Methods  & mIoU & AP$_{50}^p$  & AP$_{vol}^p$ & PCP$_{50}$  \\ \hline\hline
			\multicolumn{5}{l}{\textit{top-down models:}} \\ \hline
			P-RCNN\!~\cite{yang2019parsing} & 65.9 & 43.5 & 53.1 & 51.8  \\
			M-CE2P\!~\cite{ruan2019devil}   & 67.1 & 43.7 & 52.9 & 51.2   \\
			RP-RCNN\!~\cite{yang2020eccv}   & 65.3 & 48.5 & 54.5 & 51.1  \\
			\hline\hline
			
			\multicolumn{5}{l}{\textit{bottom-up models:}} \\ \hline
			PGN\!~\cite{gong2018instance}& 46.1 & 23.4 & 35.9 & 32.5\\
			NAN\!~\cite{zhao2018understanding} & 58.9 & 37.6 & 48.3 & 43.9  \\
			\textbf{Ours} & 69.1 & 49.7 & 54.7 & 52.8  \\	\hline
		\end{tabular}
	}
	\captionsetup{font=small}
	\caption{\small \textbf{Quantitative performance comparison on DensePose-COCO} \texttt{test}\!~\cite{alp2018densepose}, with  mIOU, AP$^p$ and PCP. See \S\ref{sec:qr} for details.}
	\label{table:densepose}
	\vspace{-14pt}
\end{table}

\noindent\textbf{Reproducibility.} Our model is implemented in PyTorch and trained on eight NVIDIA Tesla V100 GPUs with a 32GB memory per-card. Testing is conducted on a single NVIDIA Xp GPU with 11 GB memory. 

\subsection{Quantitative Results}\label{sec:qr}

\noindent\textbf{MHP$_{v2}$}~\cite{zhao2018understanding}:~Table~\ref{table:mhp} reports comparison results against five top-down and three bottom-up approaches on MHP$_{v2}$\!~\texttt{val}. Our approach significantly outperforms existing bottom-up models, \ie, MHParser\!~\cite{li2017multiple} and NAN\!~\cite{zhao2018understanding}, across all metrics. This indicates the importance of our multi-granularity representation learning in instance-level parsing. Furthermore, our approach shows better overall performance than most top-down methods (\eg, SNT\!~\cite{ji2019learning}, P-RCNN~\cite{yang2019parsing}, M-CE2P~\cite{ruan2019devil}), revealing our appealing performance. Compared to RP-RCNN~\cite{yang2020eccv}, our approach only performs slightly worse in terms of instance-level metrics (\eg, AP$_{50}^p$). However, our model improves the mIoU by $4\%$ and is much more efficient (see \S\ref{sec:runtime}).


\begin{table}
	\centering
	\small
	\resizebox{0.49\textwidth}{!}{
		\setlength\tabcolsep{12pt}
		\renewcommand\arraystretch{1.05}
		\begin{tabular}{r||cccc}
			\rowcolor{mygray}
			\thickhline
			Methods & AP$_{50}^r$ & AP$_{60}^r$ & AP$_{70}^r$ & AP$_{vol}^r$  \\ \hline\hline
			\multicolumn{5}{l}{\textit{top-down models:}} \\ \hline
			MNC\!~\cite{dai2016instance}    & 38.8 & 28.1 & 19.3 & 36.7 \\
			Li~\textit{et al.}\!~\cite{li2017holistic} & 40.6 & 30.4 & 19.1 & 38.4  \\
			HAZN\!~\cite{xia2016zoom}       & 43.7 & - & - & -\\
			P-RCNN\!~\cite{yang2019parsing} & 52.9 & 43.0 & 31.1 & 48.5 \\
			M-CE2P\!~\cite{ruan2019devil}   & 53.3 & 45.6 & 31.6 & 51.9 \\
			RP-RCNN\!~\cite{yang2020eccv}   & 59.9 & 51.3 & 37.8 & 55.8 \\
			\hline\hline
			\multicolumn{5}{l}{\textit{bottom-up models:}} \\ \hline
			PGN\!~\cite{gong2018instance}&39.6 &29.9 &20.0 &39.2 \\
			MH-Parser\!~\cite{li2017multiple} & 42.3 & 34.2 & 20.1  & 40.0 \\
			NAN\!~\cite{zhao2018understanding}& 59.7 & 51.4 & 38.0 & 52.2  \\
			\textbf{Ours} & 59.0 & 52.3 & 38.1 & 55.9 \\\hline
		\end{tabular}
	}
	\captionsetup{font=small}
	\caption{\small \textbf{Quantitative performance comparison on PASCAL-Person-Part} \texttt{test}\!~\cite{xia2017joint}, with AP$^r$. See \S\ref{sec:qr} for details.}
		\vspace{-4pt}
	\label{table:pascal}
\end{table}

\noindent\textbf{DensePose-COCO}~\cite{alp2018densepose}:
Table~\ref{table:densepose} presents comparisons with five representative approaches on DensePose-COCO~\texttt{val}. Our method sets new state-of-the-arts across all metrics, outperforming all other methods by a large margin. For example, our parser provides a considerable performance gain in AP$_{50}^p$, \ie, $1.2\%$ and $12.1\%$ higher than the current best top-down (\ie, RP-RCNN~\cite{yang2020eccv}) and bottom-up (\ie, NAN~\cite{zhao2018understanding}) models, respectively. 


\noindent\textbf{PASCAL-Person-Part}~\cite{xia2017joint}:\!~Table~\ref{table:pascal} compares our model against six top-down and three bottom-up methods on PASCAL-Person-Part \texttt{val}. The results demonstrate that our human parser outperforms other competitors across most metrics (\ie, AP$_{60}^r$, AP$_{70}^r$, AP$_{vol}^r$). We also observe that performance on this dataset has become saturated, due to its small-scale training set (\ie, only $1,\!716$ images).

\begin{figure}[t]
	\begin{minipage}[b]{0.6\linewidth}
		\centering
		\includegraphics[width=\linewidth]{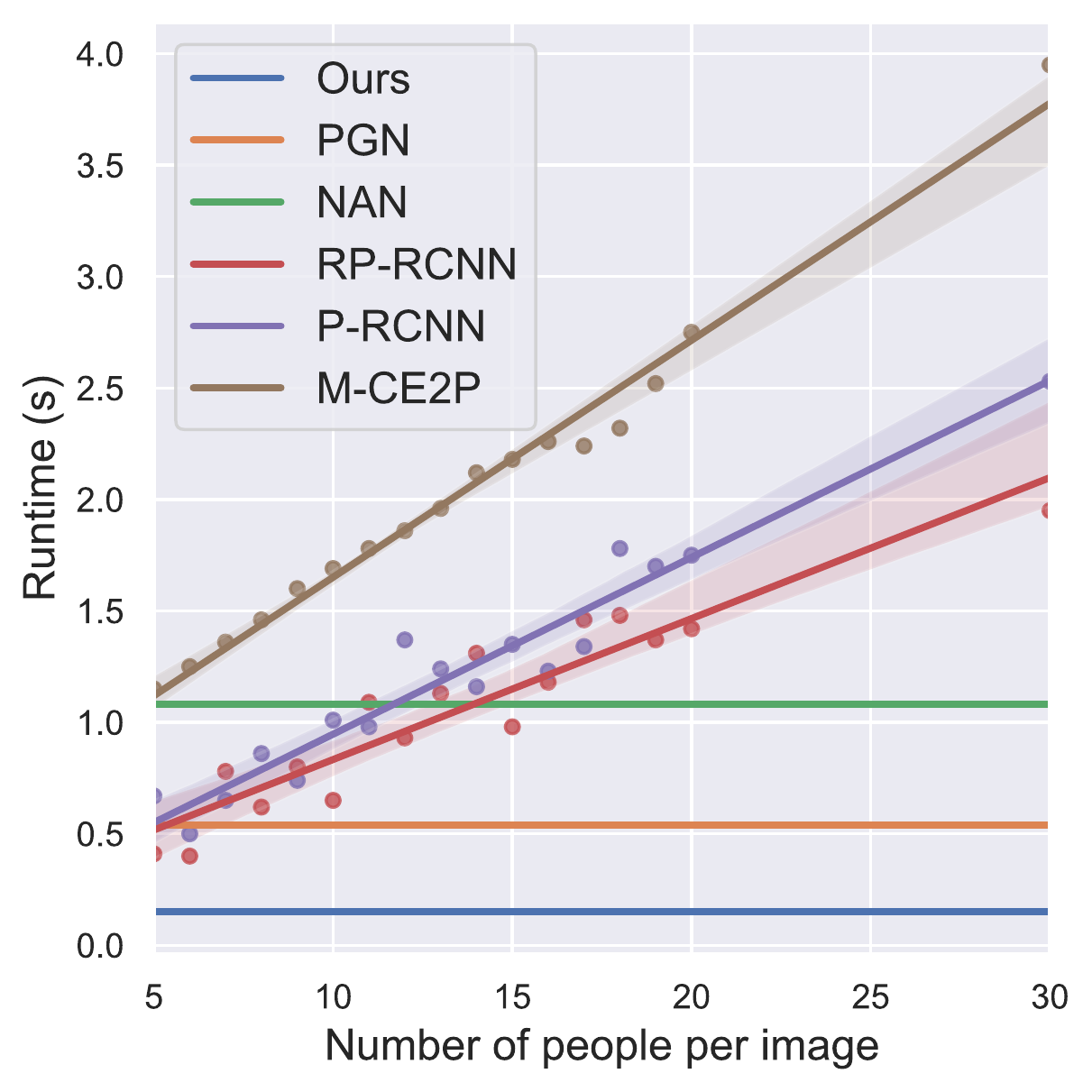}
	\end{minipage}
	\hfill\!\!\!\!\!\!\!\!
	\begin{minipage}[b]{0.43\linewidth}
		\centering
		\small
		\resizebox{\textwidth}{!}{
			\setlength\tabcolsep{1pt}
			\renewcommand\arraystretch{1.2}
			\begin{tabular}[b]{r||c}
				\thickhline
				\rowcolor{mygray}
                            & Speed\!~(s)\\ \rowcolor{mygray}
				\multirow{-2}{*}{Methods} & {\small\textit{avg. for 6 persons}}\\ \hline\hline
				\multicolumn{2}{l}{\textit{top-down models:}} \\ \hline
				SNT\!~\cite{ji2019learning} & 3.27 \\
				M-CE2P\!~\cite{ruan2019devil} & 1.15 \\
				P-RCNN\!~\cite{yang2019parsing} & 0.51  \\	
				RP-RCNN\!~\cite{yang2020eccv} & 0.43 \\ \hline\hline
				
				\multicolumn{2}{l}{\textit{bottom-up models:}} \\ \hline
				MHParser\!~\cite{li2017multiple} & 14.9 \\
				NAN\!~\cite{zhao2018understanding} & 1.08 \\
				PGN\!~\cite{gong2018instance} & 0.54 \\
				\textbf{Ours} & 0.15 \\ \hline
		\end{tabular}}
	\vspace{3pt}
	\end{minipage}
	\vspace{-18pt}
	\captionsetup{font=small}
	\caption{\small\textbf{Runtime analysis}. Our model is fully convolutional and allows for efficient inference, irrespective of the number of people in the image. In contrast, the runtimes of top-down approaches (\ie, M-CE2P\!~\cite{ruan2019devil}, P-RCNN\!~\cite{yang2019parsing}, RP-RCNN\!~\cite{yang2020eccv}) grow linearly with the number of people. The runtime comparison on MHP$_{v2}$ \texttt{val} reported in the table demonstrates again that our model is significantly faster than other methods.  See \S\ref{sec:runtime} for details.
}
		\label{fig:runtime}
		\vspace{-12pt}
\end{figure}

%

\begin{figure*}[t]
	\vspace{-5pt}
	\begin{center}
		\includegraphics[width=\linewidth]{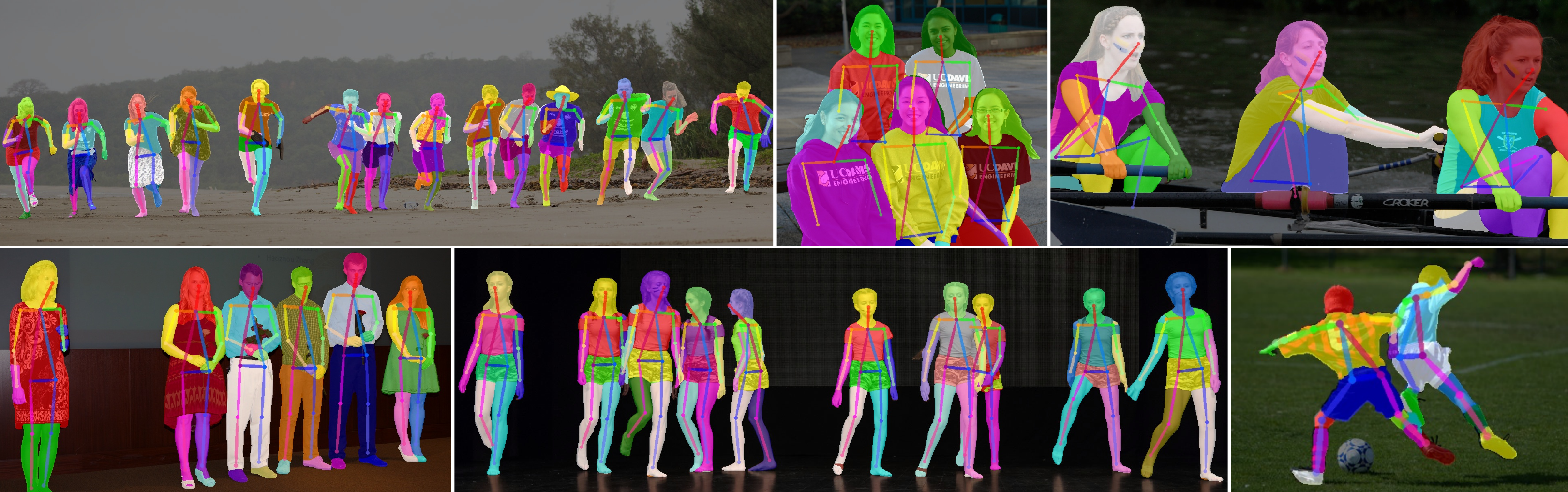}
	\end{center}
	\vspace{-17pt}
	\captionsetup{font=small}
	\caption{\small\textbf{Instance-level human semantic parsing results} (\S\ref{sec:vr}) in challenging scenarios with crowding, occlusions, pose variations, \etc.}
	\label{fig:visual}
	\vspace{-12pt}
\end{figure*}

\subsection{Runtime Analysis}\label{sec:runtime}
We collect images with a varying number of people for runtime analysis. Each analysis was repeated 500 times and then averaged. We compare with six state-of-the-art methods, including three top-down\!~(\ie, P-RCNN\!~\cite{yang2019parsing}, RP-RCNN\!~\cite{yang2020eccv}, M-CE2P\!~\cite{ruan2019devil}) and three bottom-up\!~(\ie, NAN\!~\cite{zhao2018understanding}, MHParser\!~\cite{li2017multiple}, PGN\!~\cite{gong2018instance}) models.
As depicted in\!~\figref{fig:runtime}, the inference times of the top-down models are roughly proportional to the number of people in the image, while they are invariant for the bottom-up approaches. The table summarizes the inference time of several parsers on MHP$_{v2}$\!~\texttt{val} with average six people per image. Overall, our model is much faster than existing methods.

\vspace{-1pt}
\subsection{Qualitative Results}\label{sec:vr}\vspace{-1pt}
As shown in \figref{fig:visual}, our approach can produce precise instance-level human semantic parsing results. It shows strong robustness to many real-world challenges, such as crowding, occlusions,  pose and appearance variations, \etc.

\begin{table}[t]
	\begin{centering}
	\small
	\resizebox{0.49\textwidth}{!}{
		\setlength\tabcolsep{2pt}
		\renewcommand\arraystretch{1.0}		
\begin{tabular}{ccc||ccc}
			\hline\thickhline
			\rowcolor{mygray}
			coarse-to-fine & residual refine. & learn. warping &  &   &  \\
						\rowcolor{mygray}
			(Eq.\eqref{eq:dspf}) & (Eq.\eqref{eq:res}) & (Eqs.\!~(\ref{eq:featureflow}-\ref{eq:warp})) & \multirow{-2}{*}{AP$_{50}^p$} & \multirow{-2}{*}{AP$_{vol}^p$} & \multirow{-2}{*}{PCP$_{50}$}\\
			
			\hline \hline
				&  &  & 35.5 & 40.8 & 39.3 \\
			\ding{51}	&  &  & 36.9 & 42.4 &  40.3 \\
			\ding{51}	& \ding{51} & & 38.1 & 43.4 & 41.5 \\
			\ding{51}	& \ding{51} & \ding{51} & 38.9 & 44.3  & 42.2 \\ \hline
		\end{tabular}
		
	}
	\captionsetup{font=small}
	\caption{\small\textbf{Ablation study on DSPF estimation} (\S\ref{sec:dspf}) on MHP$_{v2}$ \texttt{val}~\cite{zhao2018understanding}. See \S\ref{sec:abs} for details. }
	\label{table:dspf}
		\vspace{-10pt}
\end{centering}
\end{table}


\subsection{Diagnostic Experiments}\label{sec:abs}


\noindent\textbf{DSPF Breakdown.} Table~\ref{table:dspf} investigates the necessity of our essential modules for DSPF estimation (\S\ref{sec:dspf}). We start the analysis with a baseline model (\ie, $1_{st}$ row) which directly regresses DSPF over multi-layer features $\{\bm{X}_l\}_{l=1}^L$. Then, we progressively improve the baseline with: (1) a `coarse-to-fine' strategy (\ie, inferring a complete DSPF at each pyramid level); (2)`residual refinement' (\ie, using residual learning with bilinear feature unsampling); and (3) `learnable feature warping' (\ie, learning cross-layer feature alignment). Consistent performance improvements can be observed after introducing the above modifications and we can draw three essential conclusions: \textbf{(1)} Coarse-to-fine inference is critical for accurate DSPF regression, since it makes full use of multi-scale information. \textbf{(2)} Residual learning is also useful, since the small magnitude of the residue facilitates training. \textbf{(3)} With the learnable feature warping, cross-layer features/DSPF predictions are better aligned, further facilitating network learning.

\noindent\textbf{Differentiable Matching.} We further study the proposed differentiable solver (\S\ref{sec:jda}) for human joint association by comparing it with two greedy matching algorithms\!~\cite{papandreou2018personlab,cao2017realtime} on MHP$_{v2}$ \texttt{val} (see Table~\ref{table:pose}). The differentiable solver allows directly back-propagating joint grouping errors, leading to great improvements in multi-human pose estimation (under mAP$_{pose}$\!~\cite{lin2014microsoft}). In addition, benefiting from end-to-end learning, our model also shows better human part parsing results (in terms of AP$_{50}^p$ and mIoU), further confirming the superiority of our differentiable matching strategy.

\begin{table}[t]
	\centering
	\small
	\resizebox{0.49\textwidth}{!}{
		\setlength\tabcolsep{2pt}
		\renewcommand\arraystretch{1.05}
		\begin{tabular}{l||ccc|c|c}
			\hline\thickhline
			\rowcolor{mygray}
			Diff. Matching\!~(\S\ref{sec:jda}) & AP$_{50}^p$ & AP$_{vol}^p$   & PCP$_{50}$ & mIoU & mAP$_{pose}$ \\ \hline \hline
			greedy matching\!~\cite{papandreou2018personlab} & 37.6  & 43.1 & 40.8 & 40.8 & 63.9\\
			greedy matching\!~\cite{cao2017realtime} & 37.9  & 43.2 & 40.6 & 41.0 & 64.3\\
			differentiable matching & 39.0 & 44.3  & 42.3 & 41.4 & 65.8\\ \hline
		\end{tabular}
	}
	\captionsetup{font=small}
	\caption{\small\textbf{Ablation study on differentiable matching} (\S\ref{sec:jda}) on MHP$_{v2}$ \texttt{val}~\cite{zhao2018understanding}. See \S\ref{sec:abs} for details. }
	\label{table:pose}
		\vspace{-12pt}
\end{table}

\vspace{-2pt}
\section{Conclusion}
\vspace{-2pt}
This work presents a new perspective of exploring human structural information over multiple granularities to address instance-aware human semantic parsing. A dense-to-sparse projection field is learnt to associate human semantic parts with human keypoints. The field is progressively optimized with residual refinement to ease the optimization.
A differentiable joint matching solver is further proposed for body joint assembling.
These designs together yield an end-to-end trainable, bottom-up instance-aware human semantic parser.
With its accurate and fast computation, our parser is expected to pave the way for practical use and benefit vast quantities of human-centric applications.




{\small
\bibliographystyle{ieee_fullname}
\bibliography{egbib}
}

\end{document}